\documentclass[10pt,twocolumn,letterpaper]{article}

\usepackage{wacv}
\usepackage{times}
\usepackage{epsfig}
\usepackage{graphicx}
\usepackage{amsmath}
\usepackage{amssymb}
\usepackage{booktabs,enumitem}
\usepackage{algorithm}
\usepackage{algpseudocode,xcolor}

\newcommand{\mname}[0]{{\bf ConfMix}}
\newcommand{\caname}{C_{det}} 
\newcommand{\cbname}{C_{bbx}} 
\newcommand{\ccname}{C_{comb}} 
\newcommand{\squeezeup}{\vspace{-3mm}}

\newcommand\blfootnote[1]{%
  \begingroup
  \renewcommand\thefootnote{}\footnote{#1}%
  \addtocounter{footnote}{-1}%
  \endgroup
}
\wacvfinalcopy 

\ifwacvfinal
\usepackage[breaklinks=true,bookmarks=false]{hyperref}
\else
\usepackage[pagebackref=true,breaklinks=true,colorlinks,bookmarks=false]{hyperref}
\fi

\begin{document}

\title{ConfMix: Unsupervised Domain Adaptation for Object Detection 
\\via Confidence-based Mixing}

\author{Giulio Mattolin$^1$, Luca Zanella$^2$, Elisa Ricci$^{1,2}$, Yiming Wang$^2$\\
$^1$University of Trento, Trento, Italy \quad $^2$Fondazione Bruno Kessler, Trento, Italy\\
{\tt\small lzanella@fbk.eu}}



\maketitle
\thispagestyle{empty}

\begin{abstract}
Unsupervised Domain Adaptation (UDA) for object detection aims to adapt a model trained on a source domain to detect instances from a new target domain for which annotations are not available.
Different from traditional approaches, we propose ConfMix, the first method that introduces a sample mixing strategy based on region-level detection confidence for adaptive object detector learning. We mix the local region of the target sample that corresponds to the most confident pseudo detections with a source image, and apply an additional consistency loss term to gradually adapt towards the target data distribution. In order to robustly define a confidence score for a region, we exploit the confidence score per pseudo detection that accounts for both the detector-dependent confidence and the bounding box uncertainty.
Moreover, we propose a novel pseudo labelling scheme that progressively filters the pseudo target detections using the confidence metric that varies from a loose to strict manner along the training. We perform extensive experiments with three datasets, achieving state-of-the-art performance in two of them and approaching the supervised target model performance in the other.
Code is available at \href{https://github.com/giuliomattolin/ConfMix}{\color{magenta}https://github.com/giuliomattolin/ConfMix}\blfootnote{This work has been supported by the European Union’s Horizon 2020 research and innovation programme under grant agreement No. 957337, and the European Commission Internal Security Fund for Police under grant agreement No. ISFP-2020-AG-PROTECT-101034216-PROTECTOR.}.
\end{abstract}

\section{Introduction}
\label{sec:intro}

\begin{figure}[t!]
  \centering
  \includegraphics[width=0.8\linewidth]{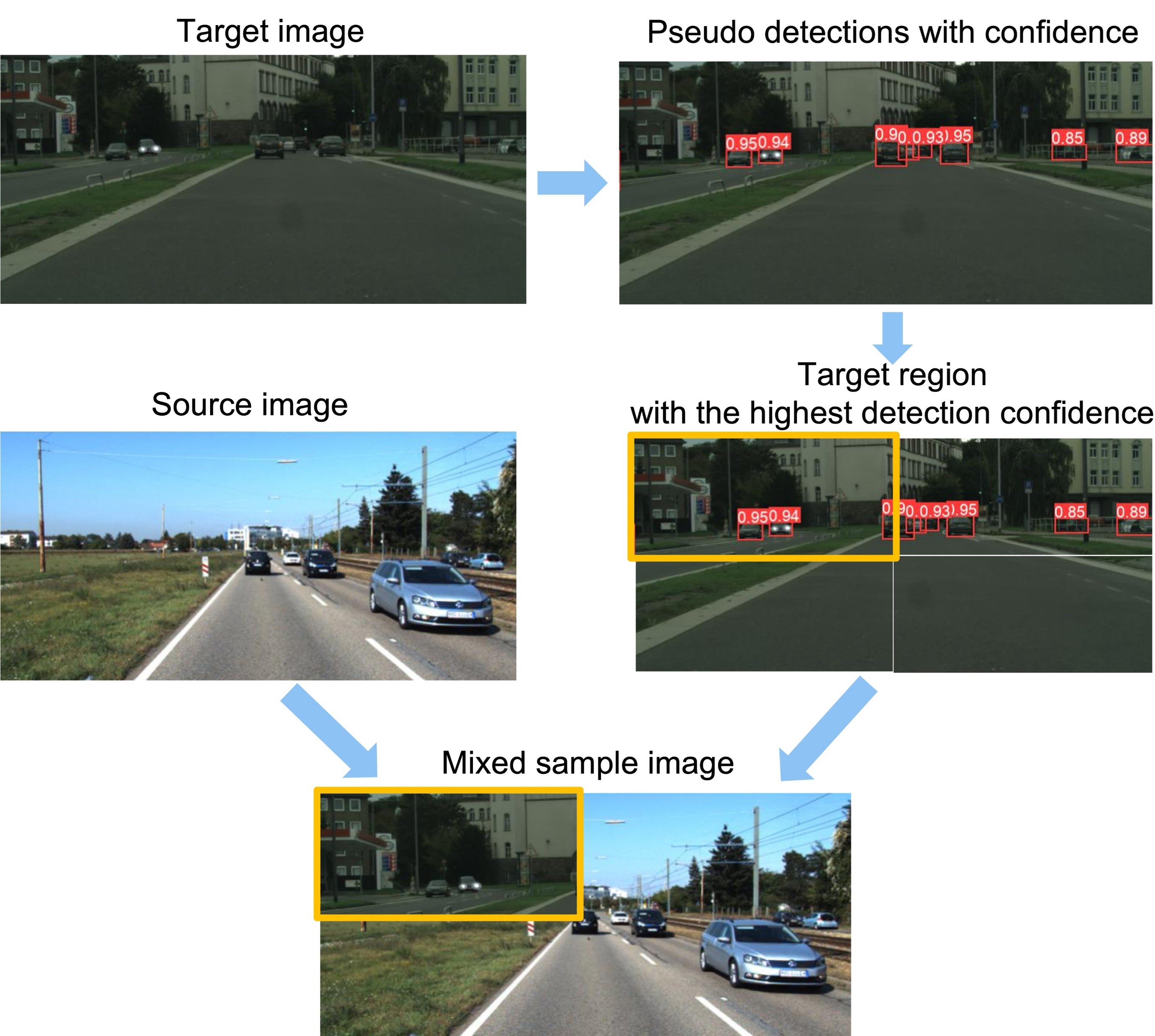}
  \caption{\mname~is based on a novel sample mixing strategy which combines the source image and the target region ({\color{orange} orange} box) with the highest pseudo detection confidence.}
  \label{fig:teaser}
  \squeezeup
\end{figure}

Object detection is a fundamental task in computer vision which involves the classification and localisation, \textit{e.g.} by bounding boxes, of objects of interest belonging to certain predefined categories. Due to its importance in many applications such as autonomous driving, video surveillance and robotic perception, object detection has received significant attention, leading to the development of several different models~\cite{rcnn, yolo, fcos, faster-rcnn}. However, as detectors mostly rely on deep learning, it is a well known fact that they suffer from severe performance degradation when being tested on images that are \textit{visually different} from the ones encountered during training, due to the \textit{domain shift}~\cite{DBLP:journals/corr/abs-1803-03243}. 

To address this problem, recent research efforts have been put on devising Unsupervised Domain Adaptation (UDA) techniques for building deep models that can adapt from an annotated source dataset to a target one without tedious manual annotations~\cite{coral, grl, pseudo_1, imagetrans_1, attn_1}. The vast majority of UDA methods for detection resort on adversarial training and on exploiting the Gradient Reversal Layer (GRL)~\cite{grl} to perform adaptation both at image-level and instance-level~\cite{DBLP:journals/corr/abs-1803-03243, DBLP:journals/corr/abs-1812-04798, Zhu_2019_CVPR, cond_domain_norm, MeGA-CDA}. 
Other approaches mostly focus on robustly producing pseudo detections in order to effectively finetune the model on the target data~\cite{SC-UDA, Collaborative, vs2022instance}. 
In general, while over the last few years several solutions have been proposed in the literature for adapting two-stage object detectors, we argue that devising UDA approaches which can also be applied to one-stage detectors would be desirable. Indeed, the latter methods are more appropriate in applications such as autonomous driving that necessitate of real-time processing and high computational efficiency.

Concurrently, recent works in computer vision have shown the benefit of adopting sophisticated data augmentation techniques by synthesising mixed samples with target and source images in order to improve generalisation ability of deep architectures~\cite{mixup, cutmix, augmix}. These methods have been considered in the context of UDA for classification~\cite{DBLP:journals/corr/abs-2007-03141, DBLP:journals/corr/abs-1905-04215}
and semantic segmentation~\cite{DBLP:journals/corr/abs-1906-01916, DBLP:journals/corr/abs-2007-07936, DBLP:journals/corr/abs-2001-04647, DBLP:journals/corr/abs-2001-03182}, demonstrating some empirical advantage. However, extending these approaches to UDA for detection is far from trivial. 

Inspired by these previous works, in this paper we propose \mname, the first mixing-based UDA approach for object detection based on the regional confidence of pseudo detections.
The main idea behind \mname~is illustrated in Figure~\ref{fig:teaser}. Specifically, we propose to artificially generate samples by combining the region of target images where the model is most confident with source images. We also introduce during training an associated consistency loss to enforce coherent predictions among generated images. 
{Our intuition is that, by combining source and target images and forming new mixed samples, we are training our model on novel, synthetically generated sample images with reliable pseudo detections and with visual appearance close to the samples of target domain, thus improving the generalisation capabilities of the detector.}
Moreover, the quality of pseudo detections plays an essential role during adaptation, and is tightly related to the confidence metric. By exploiting a stricter confidence metric, e.g. enriching the detector-dependent confidence with bounding box uncertainty~\cite{choi2019gaussian}, one can obtain more reliable pseudo detections, however with a reduced number. To mitigate this, we propose to progressively restrict the confidence metric for pseudo labelling. With a less strict confidence metric at the initial adaptation phase, we allow more pseudo detections in order to learn the representation of the target domain, while with a gradually stricter confidence metric, we aim to improve the detection accuracy with more trustworthy pseudo detections. 
We conduct extensive experiments on different datasets (Cityscapes \cite{cityscapes} $\rightarrow$ FoggyCityscapes \cite{foggy_cityscapes}, Sim10K \cite{sim10k} $\rightarrow$ Cityscapes and KITTI \cite{kitti} $\rightarrow$ Cityscapes) and we show that our approach outperforms existing algorithms in most setups. 

We summarise our main contributions as below:
\begin{itemize}[noitemsep,nolistsep]
    \item {We introduce the first sample-mixing UDA method for object detection. Our approach, named \mname, mixes samples from source and target domains based on the regional confidence of target pseudo detections}. 
    \item {We propose a novel Progressive Pseudo-labelling scheme by gradually restricting the confidence metric along the adaptive learning, which allows for a smooth transition when learning target representation, thus improving detection accuracy}.
    \item \mname~scores the new state-of-the-art adaptation performance, achieving +1.7\% on Sim10k $\rightarrow$ Cityscapes, and +3.7\% on KITTI $\rightarrow$ Cityscapes in terms of mean Average Precision (mAP). 
\end{itemize}

\section{Related work}
\label{sec:sota}

\noindent\textbf{Object Detection.}
Current object detection models can be grouped into two main categories: one-stage and two-stage approaches. One-stage object detectors, such as YOLO \cite{yolo} and FCOS \cite{fcos}, adopt a unified framework to obtain final results directly from the feature maps generated by a CNN backbone. These frameworks are very computationally efficient and are able to achieve near real-time speed during inference. On the other hand, two-stage object detectors, such as RCNN \cite{rcnn}, generate predictions by first extracting region proposals and then, leveraging this information, produce classification labels and bounding box coordinates. Such models are widely adopted for their high performance but, although research has been conducted to improve detection speed \cite{fast-rcnn, faster-rcnn, r-fcn}, they are considerably slower compared to one-stage detectors.

\noindent\textbf{Unsupervised Domain Adaptation.}
Given a labelled source domain and an unlabelled target domain, UDA aims to use the available data to produce a model that is able to generalise and perform well on the target domain. 
A conventional approach is to reduce the domain gap by directly minimising the distance between feature distributions using discrepancy loss functions \cite{mmd,coral}.
On the other hand, adversarial-based methods \cite{grl,ganin2016domain,adda}, employ a domain discriminator and a feature extractor that learns to produce domain-invariant feature representations by fooling the discriminator. Many works demonstrated the benefit of using pseudo labels to maximally leverage information from the target domain \cite{liang2021domain,li2021cross,li2021ecacl}, eventually considering a gradual scheme for incorporating them \cite{yoon2022semi}. Other works have focused on adopting sample mixing techniques, 
such as mixup \cite{mixup} or CutMix \cite{cutmix}, to improve generalisation. 
For instance, in \cite{DBLP:journals/corr/abs-2007-03141,xu2020adversarial} domain-level mixup regularisation is applied to ensure domain invariance in the learned feature representations, while in \cite{chen2022transmix,park2022saliency} the model's attention is used to re-assign the confidence of saliency-guided samples and labels. 
Similar ideas are implemented in previous works considering the segmentation task \cite{DBLP:journals/corr/abs-1906-01916, DBLP:journals/corr/abs-2007-07936, DBLP:journals/corr/abs-2001-04647, DBLP:journals/corr/abs-2001-03182, DBLP:journals/corr/abs-2105-08128}. However, to the best of our knowledge, no previous works have been proposed to exploit mixing techniques for UDA in the context of object detection. 


\noindent\textbf{UDA for Object Detection.}
In the context of object detection, UDA was recently introduced by \cite{DBLP:journals/corr/abs-1803-03243}, which proposed image- and instance-level alignment using two GRLs \cite{grl} on Faster R-CNN. Subsequently, several methods started to address this problem mainly using two-stage detectors. Focusing on image-level, \cite{DBLP:journals/corr/abs-1812-04798} showed that strong-local alignment and weak-global alignment of the features extracted from the backbone improve adaptation, while \cite{Zhu_2019_CVPR}, focusing on instance-level, exploited RPN proposals to perform region-level alignment. To adapt the source-biased decision boundary to the target data, \cite{DBLP:journals/corr/abs-2003-06297} combined adversarial training with image-to-image translation by generating interpolated samples using Cycle-GAN \cite{cyclegan}. Other recent works have proposed applying self-training with pseudo detections to perform the adaptation. To address the risk of performance degradation caused by overfitting noisy pseudo detections, \cite{SC-UDA} introduces an uncertainty-based fusion of pseudo detections sets generated via stochastic inference, \cite{Free_Lunch} proposes self-entropy descent (SED) as a metric to search for an appropriate confidence threshold for reliable pseudo detections, while \cite{vs2022instance} uses a student-teacher framework and gradually updates the source-trained model. 

Few works have addressed UDA for one-stage detectors, e.g. FCOS \cite{SCAN, li2022sigma, everypixelmatters} or SSD \cite{selftraining_onestage}. In particular, adopting a self-training procedure reduces the negative effects of inaccurate pseudo detections by performing hard negative pseudo detections mining followed by a weak negative mining strategy, where instance-level scores are computed for each detection considering all neighbouring boxes~\cite{selftraining_onestage}. 
In addition, adversarial learning is employed using GRL~\cite{grl} and a discriminator with the aim of extracting discriminative background features and reducing the domain shift. However, our approach is radically different, as it does not require additional architectural components to the network, but proposes a mixing-based data augmentation strategy to promote regularisation of the model. 

\section{Method}
\label{sec:method}

\begin{figure*}[t!]
  \centering
  \includegraphics[width=0.8\linewidth]{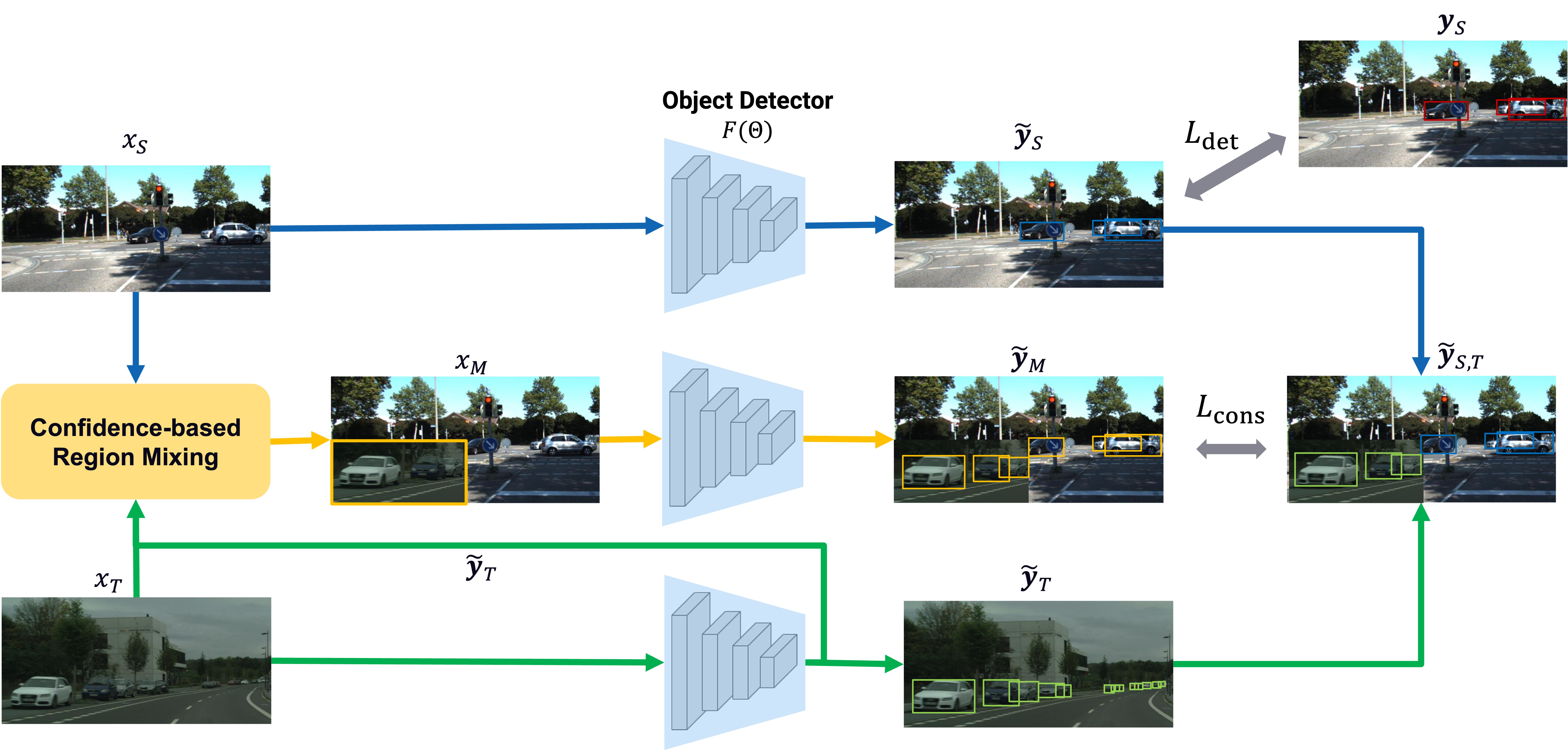}
  \caption{Overview of the proposed \mname~method. We pass the source sample $x_S$ and target sample $x_T$ to the detector model $F(\Theta)$, obtaining the prediction $\tilde{\textbf{y}}_S$ and $\tilde{\textbf{y}}_T$, respectively. We select the target region with the highest regional confidence to form the mixed sample $x_M$ with the source image, which is then fed to the detector model $F(\Theta)$, producing predictions $\tilde{\textbf{y}}_M$. We train the model with the supervised detection loss using the source annotations $\textbf{y}_S$ and the self-supervised consistency loss by comparing $\tilde{\textbf{y}}_M$ with the combined source and target predictions $\tilde{\textbf{y}}_{S,T}$.}
  \label{fig:method}
  \squeezeup
\end{figure*}

The proposed \mname, as illustrated in Figure \ref{fig:method}, synthesises an image $x_M \in R^{W \times H \times C}$ by mixing a source image $x_S \in R^{W \times H \times C}$ and the local region of a target image $x_T \in R^{W \times H \times C}$ with the most reliable pseudo detections. We first predict a set of $N_T$ pseudo detections ${\tilde{\textbf{y}}_{T}} = \left\{\tilde{{y}}_{T}^i | i\in [1,~N_T]\right\}$ on the target image and compute the confidence per pseudo detection using the detector network $F(\Theta)$ that is parameterised with $\Theta$ and is originally trained only on the source data. We opt to follow a Gaussian modelling of the bounding box predictions, instead of the deterministic one, in order to improve the reliability of the detector confidence with the uncertainty of the bounding box prediction.  
Next, we divide the target image $x_T$ into regions of equal size and select the region with the highest average confidence of pseudo detections to mix with the source sample $x_S$, forming the mixed sample $x_M$.

We pass $x_T$, $x_S$, and $x_M$ to the detector $F(\Theta)$ and obtain their corresponding detections $\tilde{\textbf{y}}_{T}$, $\tilde{\textbf{y}}_{S}$ and $\tilde{\textbf{y}}_{M}$, respectively. The detector then learns to adapt to the target domain by imposing a consistency loss $L_{cons}$ which promotes the similarity between $\tilde{\textbf{y}}_{M}$ and the combined detections $\tilde{\textbf{y}}_{S,T}$ by merging the source $\tilde{\textbf{y}}_{S}$ and target $\tilde{\textbf{y}}_{T}$ detections according to how the two sample images are mixed. The supervision of source ground-truth detections $\textbf{y}_{S}$ is achieved with the detector-related loss $L_{det}$ in order to maintain the detector capability during adaptation.

In the following sections, we describe our proposed \mname~in details, where we first introduce the estimation of the Gaussian-based detection confidence in Sec.~\ref{sec:method:confidence}, followed by the confidence-based region mixing strategy for synthesising training samples in Sec.~\ref{sec:method:mixing} and the progressive pseudo labelling in Sec. \ref{sec:method:adaptation}. Finally, we present the training objectives with losses in Sec.~\ref{sec:method:loss_training}.

\subsection{Gaussian-based detection confidence}
\label{sec:method:confidence}

Conventional object detectors, such as YOLO \cite{yolo}, Faster R-CNN \cite{faster-rcnn} and FCOS \cite{fcos}, compute and assign to each detection a confidence score $\caname \in [0,~1]$ that is often detector-dependent and is used to filter out unreliable predictions via non-maximum suppression. 
However, such confidence score does not account for the reliability of the predicted bounding box $\mathbf{b} = \left[ b_x,~b_y,~b_h,~b_w \right]$, where $\left [b_x,~b_y\right ]$ are the position of bounding box on the image and $b_h$ and $b_w$ represent the height and width, respectively. 
As suggested in~\cite{choi2019gaussian}, by taking into consideration both the detector-dependent confidence and the confidence that is derived from the uncertainty of bounding box prediction, one can improve the reliability of pseudo detections and reduce the number of false positives.

In order to compute the bounding box uncertainty, $\mathbf{b}$ requires a Gaussian-based modelling. Specifically, for each element in $\mathbf{b}$, the detector model predicts both a mean $\mu$ and a variance $\Sigma$, where the variance represents the localisation uncertainty. 
Thus, we can express the Gaussian-based bounding box $\hat{\mathbf{b}}$ as:
\begin{equation}
    \hat{\mathbf{b}} = \left[{\mu}_{bx},~{\mu}_{by},~{\mu}_{bh},~{\mu}_{bw},~\Sigma_{bx},\Sigma_{by},\Sigma_{bh},\Sigma_{bw}\right],
\end{equation}
where both the means $\hat{\mathbf{b}}_{\mu} = \left[{\mu}_{bx},~{\mu}_{by},~{\mu}_{bh},~{\mu}_{bw}\right]$ and the variances $\hat{\mathbf{b}}_{\Sigma} = \left[~\Sigma_{bx},\Sigma_{by},\Sigma_{bh},\Sigma_{bw}\right]$ are predicted by the detector with an updated regression loss (see details in Sec.~\ref{sec:method:loss_training}). Note that a sigmoid function $\sigma(\cdot)$ is applied to the predicted variance value to ensure its range is between 0 and 1. 

As a larger variance value implies a higher uncertainty, the confidence of a bounding box is computed as:

\begin{equation}
    \cbname = 1 - mean(\hat{\mathbf{b}}_{\Sigma}),
\label{eq:bbox_unc}
\end{equation}
where $mean(\cdot)$ computes the average variance of $\hat{\mathbf{b}}_{\Sigma}$.

The combined confidence can thus be computed as:
\begin{equation}
    \ccname = \caname \cdot \cbname.
\label{eq:cal_conf}
\end{equation}

\subsection{Confidence-based Region Mixing}
\label{sec:method:mixing}
With the estimated confidence for each pseudo detection on the target image, we design a novel mixing strategy to synthesise new training samples with highly reliable pseudo detections. Instead of extracting only pseudo detections or randomly selecting part of the target image~\cite{cutmix} to mix, we propose a novel region-level mixing strategy whose synthesised samples contain both the foreground and background features from the two domains, contributing to a more effective adaptation towards the target domain. 

Specifically, we randomly sample a source image $x_S$ and a target image $x_T$. The target $x_T$ is then passed to the object detector $F(\Theta)$, producing the predictions $\tilde{\textbf{y}}_T$. 
The target image $x_T$ is then evenly divided into 4 regions as shown in Figure~\ref{fig:teaser}. Each region is considered to contain a prediction $\tilde{y}_T^i$ if its centre coordinate resides within the region. The region confidence is computed as the average of the confidences of all the pseudo detections that lie within the region. We select the region with the highest region confidence to mix with the sampled source image $x_S$ and generate the synthesised image $x_M$:

\begin{equation}
    x_M = M_T \odot x_S + (1-M_T) \odot x_T,
    \label{eq:mix_image}
\end{equation}
where $M_T \in R^{W \times H}$ is the mask matrix indicating which pixels of the target image should be masked.














\subsection{Progressive Pseudo Labelling}
\label{sec:method:adaptation}

The correctness of the pseudo detections are tightly related to the confidence metric which is used for filtering the detections. At the early stage of the adaptation, the confidence tends to be less reliable and in general of a lower value due to the large domain gap. Thus, the non-maximum suppression ends up in filtering out most of the pseudo detections if a strict confidence metric, such as $\ccname$, is applied. Therefore, we propose to perform a gradual transition of the confidence metric from a loose to strict manner, to first learn an initial representation of the target domain by allowing more pseudo detections and then gradually shift towards a stricter confidence metric to improve detection accuracy with more reliable pseudo detections.


To this end, we start with the loose confidence metric $\caname$ for filtering the pseudo detections. As iterations continue, we progressively assign more importance to $\ccname$ with a shifting weight $\delta$: 
\begin{equation}
    C = (1 - \delta) \cdot \caname + \delta \cdot \ccname.
\label{eq:confidence}
\end{equation}

The shifting weight $\delta$ varies based on the progress of the training, thus it is dependent on the iteration $t$, epoch $e$ and the number of batches in one epoch $N_b$. We devise $\delta$ with a non-linear function to gradually increase from 0 to 1:
\begin{align}
    \delta &= \frac{2}{1 + exp(-\alpha \cdot r)} - 1, \label{eq:delta}\\
    r &= \frac{t}{N_b \cdot e}, \label{eq:ratio}
\end{align}
where $r$ is the ratio of the current iteration to the total number of iterations, with its scale modulated by $\alpha$.

The pseudo detection with a confidence value that is higher than a predefined threshold value $C_{th}$, i.e. $C>C_{th}$, is considered a valid detection, and it will be accounted during the confidence-based region mixing and the training for detector adaptation. 

\subsection{Adaptive detector training}
\label{sec:method:loss_training}

To facilitate the adaptive learning of the detector $F(\Theta)$, we rely on two main losses: a self-supervised consistency loss term $L_{cons}$ on the mixed samples and a supervised detector loss term $L_{det}$ on the labelled source samples. 
$L_{det}$ aims to maintain the task-specific knowledge during adaptation, while the consistency loss $L_{cons}$ aims to adapt towards the target representation by penalising the difference between $\tilde{\textbf{y}}_{M}$ and the combined detections $\tilde{\textbf{y}}_{S,T}$ by merging the source $\tilde{\textbf{y}}_{S}$ and target $\tilde{\textbf{y}}_{T}$ detections based on how the mixed samples are formed. 

Specifically, let $\tilde{\textbf{y}}_T^R$ be the set of target pseudo detections residing within the selected target region, while $\tilde{\textbf{y}}_S^{R-}$ be the set of source pseudo detections residing outside the selected target region. The combined detections $\tilde{\textbf{y}}_{S,T}$ is the union of the two sets, i.e. $\tilde{\textbf{y}}_{S,T} = \left\{\tilde{\textbf{y}}_T^R,~\tilde{\textbf{y}}_S^{R-}\right\}$. It can happen that the bounding box dimension of $\tilde{y}_T^i \in \tilde{\textbf{y}}_T^R$ (or $\tilde{y}_S^i \in \tilde{\textbf{y}}_S^{R-}$) can exceed the selected target (or source) region, leading to inaccurate pseudo detections. We therefore clip such bounding boxes by their corresponding region boundary. 

We define $L_{cons}$ and $L_{det}$ as  $L_{cons} = \mathcal{L}(\tilde{\textbf{y}}_{M},~\tilde{\textbf{y}}_{S,T})$ and $L_{det} = \mathcal{L}(\tilde{\textbf{y}}_{S},~\textbf{y}_S)$,
where both the supervised detection loss $L_{det}$ and the self-supervised consistency loss $L_{cons}$ share the same loss function $\mathcal{L}(\cdot)$. While $L_{det}$ aims to penalise the difference between the predicted detections $\tilde{\textbf{y}}_S$ and the ground-truth detections $\textbf{y}_S$ on the source samples, $L_{cons}$ aims to penalise the difference between the predicted detections $\tilde{\textbf{y}}_{M}$ and the pseudo detections $\tilde{\textbf{y}}_{S,T}$ on the mixed samples. 
Note that $\mathcal{L}(\cdot)$ is dependent on the employed object detector. In the case of the one-stage YOLOv5, $\mathcal{L}(\cdot)$ is a combination of three terms: $L_{box}$ is a Complete-IoU (CIoU) loss for regressing the bounding box coordinates, $L_{obj}$ is the Binary Cross Entropy (BCE) loss for the objectness score and $L_{cl}$ is a BCE loss for the classification score. 


In particular, as our predicted bounding box follows a Gaussian modelling, the regression loss per sample image is updated as follows:
\begin{equation}
    L_{box} = \frac{1}{N}\sum_{i=1}^{N}(1 - mean(\mathcal{N}(y^i | \hat{\textbf{b}}_{\mu}^i, \hat{\textbf{b}}_{\Sigma}^i)),
    \label{eq:reg_loss}
\end{equation}
where $\mathcal{N}(\cdot)$ is the probability density function of a normal distribution for calculating the conditional probability of obtaining the ground-truth $y^i\in\textbf{y}_S$ for $L_{det}$, or pseudo detection $y^i\in\tilde{\textbf{y}}_{S,T}$ for $L_{cons}$, given the respective means $\hat{b}_{\mu}^i$ and variances $\hat{b}_{\Sigma}^i$ predicted by the object detector. $N$ represents the total number of $\left\{y^i\right\}$.

Finally, the total loss is expressed as a weighted sum of $L_{det}$ and $L_{cons}$: 
\begin{equation}
    L_{total} = L_{det} + \gamma L_{cons},
    \label{eq:loss_function}
\end{equation}
where $\gamma$ is a hyperparameter to balance the supervised and self-supervised terms. The consistency loss $L_{cons}$ can have a greater importance when the pseudo detections are more reliable, and vice versa. We therefore define $\gamma$ as the ratio of the number of pseudo detections on $x_M$ with confidence greater than $C_{th}^{\gamma}$ and the total number of pseudo detections after non-maximum suppression, to reflect the reliability of the pseudo detections: 
\begin{equation}
    \gamma = \frac{\left|\left\{ \tilde{y}_{S,T}^i \in \tilde{\textbf{y}}_{S,T}^i: C^i \geq C_{th}^{\gamma}\right\}\right|}{|\tilde{\textbf{y}}_{S,T}|},
    \label{eq:cons_weight}
\end{equation}
where $|\cdot|$ is the cardinality of a set.


\section{Experiments}
\label{sec:exp}
We evaluate our proposed method \mname~against state-of-the-art methods on three common benchmark adaptation scenarios, together with extensive ablation studies to prove the effectiveness of our design choices.

\noindent\textbf{Datasets.} We evaluate our method 
on the four datasets:
\begin{itemize}[noitemsep,nolistsep]
    \item \textbf{Cityscapes \cite{cityscapes}} is a collection of urban street scenes for semantic understanding. Images were collected in 50 cities over several months, during the day, and in good weather conditions. Single instance annotations are available for the following 8 categories: person, rider, car, truck, bus, train, motorcycle and bicycle.

    \item \textbf{FoggyCityscapes \cite{foggy_cityscapes}} is an extension of Cityscapes in which images are augmented by applying a fog filter. FoggyCityscapes includes the same images and 8 categories as Cityscapes.

    \item \textbf{Sim10K \cite{sim10k}} is a synthetic dataset consisting of 10,000 images derived from the video game Grand Theft Auto V, including only the car category.

    \item \textbf{KITTI \cite{kitti}} is a dataset of several hours of traffic video recorded by high-resolution colour and greyscale cameras, containing 7481 training images with annotations provided for 8 categories: car, van, truck, pedestrian, person sitting, cyclist, tram and misc. 
\end{itemize}

Following \cite{everypixelmatters}, we experiment on the benchmark Cityscapes $\rightarrow$ FoggyCityscapes regarding weather adaptation, Sim10K $\rightarrow$ Cityscapes regarding synthetic-to-real adaptation, and KITTI $\rightarrow$ Cityscapes regarding cross-camera adaptation. In the latter synthetic-to-real and cross-camera adaptations, we only consider the car category, while for Cityscapes $\rightarrow$ FoggyCityscapes, we consider the complete 8 categories.

\noindent\textbf{Evaluation Metrics.} We evaluate our proposed method on the target domain in terms of Average Precision (AP), which is computed by combining precision and recall for each object category separately. We obtain the mean AP (mAP) by averaging the AP across all object categories.

\noindent\textbf{Implementation details.} We based our experiments on the YOLOv5s architecture for its lightness among the YOLOv5 series,
using PyTorch and the default 
settings. We set the batch size to 2, with each batch containing a source image and a target image of the size of $600\times600$. In all our experiments, we pre-train the model on the source domain for 20 epochs with the COCO-pretrained weights as initialisation, and perform adaptive learning for 50 epochs.
In the non-maximum suppression stage, we set the IoU threshold to 0.5, and the confidence threshold $C_{th}$ to 0.25 for producing pseudo detections. For the computation of $\gamma$, we set the confidence threshold $C_{th}^{\gamma}$ to 0.5. Please refer details regarding the hyper-parameters in the Supplementary Material.

\subsection{Comparisons}
\label{sec:exp:comparisons}
We compare \mname~against recent state-of-the-art UDA approaches for adaptive object detection on three benchmarks. In particular, we compare with adversarial feature learning methods, such as MGA~\cite{zhou2022multi}, MeGA-CDA~\cite{MeGA-CDA}, SSOD~\cite{Seek_Sim}, EPM~\cite{everypixelmatters}, CDN~\cite{cond_domain_norm}, SAPN~\cite{Spatial_Attn}; pseudo-label based self-training techniques such as SC-UDA~\cite{SC-UDA}, IRL~\cite{vs2022instance}, FL-UDA~\cite{Free_Lunch}, CTRP~\cite{Collaborative}; and graph reasoning works such as SCAN~\cite{SCAN}, SIGMA~\cite{li2022sigma}, GIPA~\cite{Graph-induced}.
We also include ``Source only'', the detector model that is trained only with labelled source data, serving as the performance lower-bound, and ``Oracle'', the detector model that is trained with labelled target data, serving as the performance upper-bound. 

\noindent\textbf{Result Discussion.} Table~\ref{tab:sk2c} reports the results of \mname~and all compared methods in the synthetic-to-real scenario Sim10k $\rightarrow$ Cityscapes and the cross-camera scenario KITTI $\rightarrow$ Cityscapes. On both benchmarks, our \mname~scores the new state-of-the-art adaptation performance, achieving +1.7\% on Sim10k $\rightarrow$ Cityscapes, and +3.7\% on KITTI $\rightarrow$ Cityscapes in terms of mAP.

Table~\ref{tab:c2f} reports the per-class detection performance of \mname~and all compared methods in the weather adaptation scenario Cityscapes $\rightarrow$ FoggyCityscapes. 
The gap between \mname~and our upper-bound ``Oracle" is rather narrow, i.e. -1.5\%. Limited by the low ``Oracle" performance on this benchmark, which is lower than MGA~\cite{zhou2022multi}, it is quite unlikely our method can outperform existing approaches regarding mAP. 
Nevertheless, our method achieves a +1\% AP gain in the person class compared to SIGMA \cite{li2022sigma} and a +2\% AP gain in the car class compared to MGA \cite{zhou2022multi}, which are the approaches obtaining the second best AP in these classes, respectively. 

\begin{table}[h]
\centering
\resizebox{\linewidth}{!}{
\begin{tabular}{ccccc}
\hline
 & & & Sim10K$\rightarrow$  & KITTI$\rightarrow$  \\
 & & & Cityscapes & Cityscapes \\
\hline
Method & Detector & Backbone & mAP & mAP \\
\hline
Source only & YOLOv5 & CSP-Darknet53 & 49.5 & 39.9 \\
\hline
SC-UDA \cite{SC-UDA} & Faster R-CNN & VGG-16 & 52.4 & 46.4 \\
MeGA-CDA \cite{MeGA-CDA} & Faster R-CNN & VGG-16 & 44.8 & 43.0 \\
FL-UDA \cite{Free_Lunch} & Faster R-CNN & VGG-16 & 43.1 & 44.6 \\
CDN \cite{cond_domain_norm} & Faster R-CNN & VGG-16 & 49.3 & 44.9 \\
SAPN \cite{Spatial_Attn} & Faster R-CNN & VGG-16 & 44.9 & 43.4 \\
CTRP \cite{Collaborative} & Faster R-CNN & VGG-16 & 44.5 & 43.6 \\
IRGG \cite{vs2022instance} & Faster R-CNN & ResNet-50 & 43.2 & 45.7 \\
SSOD \cite{Seek_Sim} & Faster R-CNN & ResNet-50 & 49.3 & 47.6 \\
GIPA \cite{Graph-induced} & Faster R-CNN & ResNet-50 & 47.6 & 47.9 \\
\hline
MGA \cite{zhou2022multi} & FCOS & VGG-16 & 54.6 & 48.5 \\
SIGMA \cite{li2022sigma} & FCOS & VGG-16 & 53.7 & 45.8 \\
SCAN \cite{SCAN} & FCOS & VGG-16 & 52.6 & 45.8 \\
EPM \cite{everypixelmatters} & FCOS & ResNet-101 & 51.2 & 45.0 \\
\mname~(Ours) & YOLOv5 & CSP-Darknet53 & \textbf{56.3} & \textbf{52.2} \\
\hline
Oracle & YOLOv5 & CSP-Darknet53 & 70.3 & 70.3 \\
\hline
\end{tabular}
}
\caption{Quantitative results (mAP) for Sim10K/KITTI $\rightarrow$ Cityscapes benchmark.}
\label{tab:sk2c}
\squeezeup
\end{table}

\begin{table*}[t!]
\centering
\resizebox{\linewidth}{!}{
\begin{tabular}{cccccccccccc}
\hline
Method & Detector & Backbone & person & rider & car & truck & bus & train & motorcycle & bicycle & mAP \\
\hline
Source only & YOLOv5 & CSP-Darknet53 & 34.8 & 37.6 & 48.7 & 14.3 & 30.1 & 8.8 & 14.6 & 28.1 & 27.1 \\
\hline
SC-UDA \cite{SC-UDA} & Faster R-CNN & VGG-16 & 38.5 & 43.7 & 56.0 & 27.1 & 43.8 & 29.7 & 31.2 & 39.5 & 38.7 \\
MeGA-CDA \cite{MeGA-CDA} & Faster R-CNN & VGG-16 & 37.7 & 49.0 & 52.4 & 25.4 & 49.2 & 46.9 & 34.5 & 39.0 & 41.8 \\
FL-UDA \cite{Free_Lunch} & Faster R-CNN & VGG-16 & 30.4 & \textbf{51.9} & 44.4 & \textbf{34.1} & 25.7 & 30.3 & 37.2 & 41.8 & 37.0 \\
CDN \cite{cond_domain_norm} & Faster R-CNN & VGG-16 & 35.8 & 45.7 & 50.9 & 30.1 & 42.5 & 29.8 & 30.8 & 36.5 & 36.6 \\
SAPN \cite{Spatial_Attn} & Faster R-CNN & VGG-16 & 40.8 & 46.7 & 59.8 & 24.3 & 46.8 & 37.5 & 30.4 & 40.7 & 40.9 \\
CTRP \cite{Collaborative} & Faster R-CNN & VGG-16 & 32.7 & 44.4 & 50.1 & 21.7 & 45.6 & 25.4 & 30.1 & 36.8 & 35.9 \\
MGA \cite{zhou2022multi} & Faster R-CNN & VGG-16 & 43.9 & 49.6 & 60.6 & 29.6 & \textbf{50.7} & 39.0 & \textbf{38.3} & \textbf{42.8} & \textbf{44.3} \\
IRGG \cite{vs2022instance} & Faster R-CNN & ResNet-50 & 37.4 & 45.2 & 51.9 & 24.4 & 39.6 & 25.2 & 31.5 & 41.6 & 37.1 \\
SSOD \cite{Seek_Sim} & Faster R-CNN & ResNet-50 & 38.8 & 45.9 & 57.2 & 29.9 & 50.2 & \textbf{51.9} & 31.9 & 40.9 & 43.3 \\
GIPA \cite{Graph-induced} & Faster R-CNN & ResNet-50 & 32.9 & 46.7 & 54.1 & 24.7 & 45.7 & 41.1 & 32.4 & 38.7 & 39.5
 \\
\hline
SCAN \cite{SCAN} & FCOS & VGG-16 & 41.7 & 43.9 & 57.3 & 28.7 & 48.6 & 48.7 & 31.0 & 37.3 & 42.1 \\
SIGMA \cite{li2022sigma} & FCOS & ResNet-50 & 44.0 & 43.9 & 60.3 & 31.6 & 50.4 & 51.5 & 31.7 & 40.6 & 44.2 \\
EPM \cite{everypixelmatters} & FCOS & ResNet-101 & 41.5 & 43.6 & 57.1 & 29.4 & 44.9 & 39.7 & 29.0 & 36.1 & 40.2 \\
\mname~(Ours) & YOLOv5 & CSP-Darknet53 & \textbf{45.0} & 43.4 & \textbf{62.6} & 27.3 & 45.8 & 40.0 & 28.6 & 33.5 & 40.8 \\
\hline
Oracle & YOLOv5 & CSP-Darknet53 & 42.3 & 43.9 & 65.9 & 33.6 & 45.0 & 37.5 & 29.3 & 36.7 & 42.3 \\
\hline
\end{tabular}
}
\caption{Quantitative results (mAP) for Cityscapes $\rightarrow$ Foggy Cityscapes benchmark.}
\label{tab:c2f}
\squeezeup
\end{table*}

\noindent\textbf{Real-time analysis.} We perform all experiments on a single NVIDIA Tesla V100. At adaptive training time, each epoch took $\sim$13 minutes with a batch size of 2, while training YOLOv5 per epoch the time is $\sim$6 minutes on Sim10K $\rightarrow$ Cityscapes. On the Cityscape dataset, the detection speed of \mname~is 76 frames per second, almost equal to the 79 frames per second of YOLOv5.

\begin{figure*}[t!]
  \centering
  \includegraphics[width=0.6\linewidth]{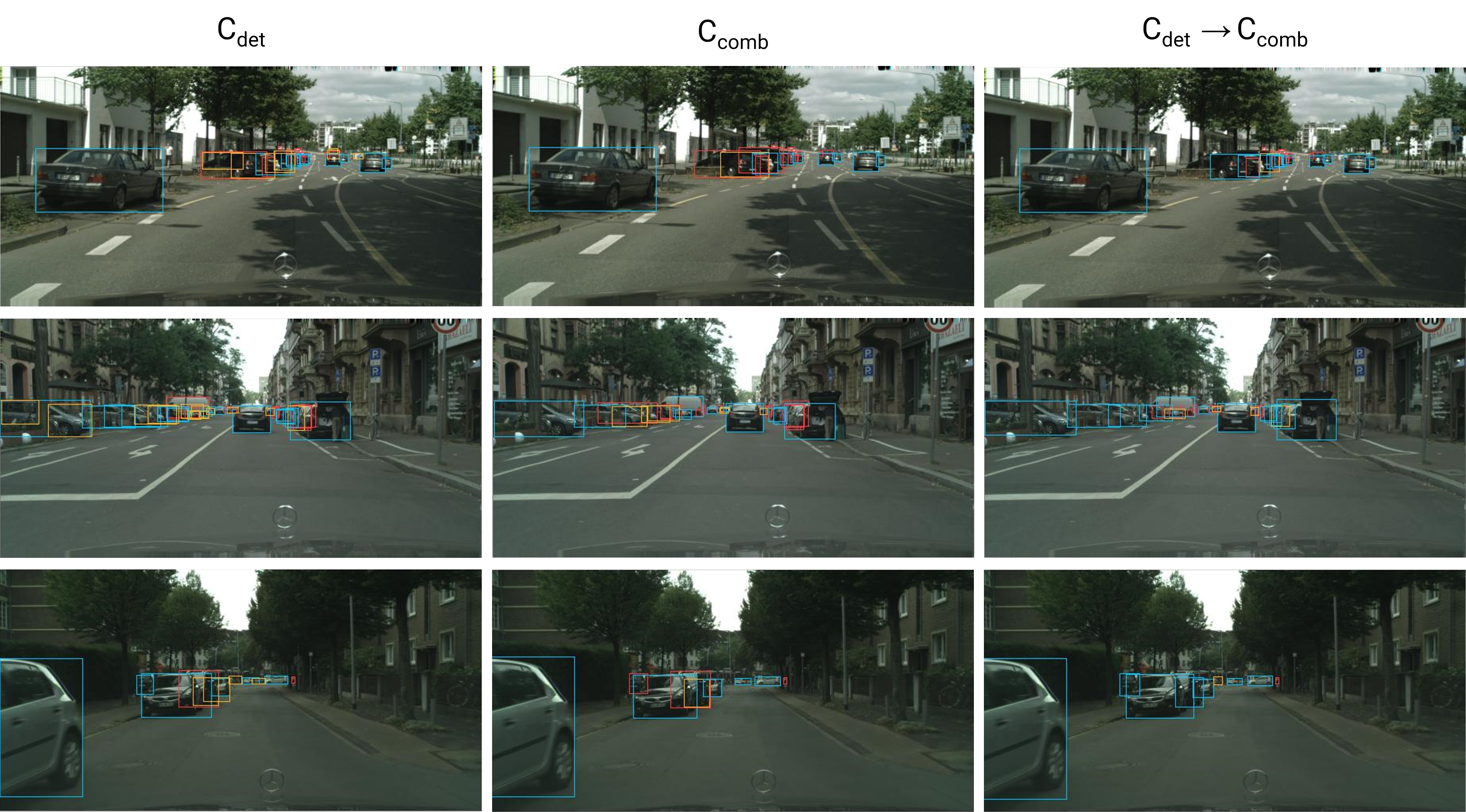}
  \caption{Qualitative results on Sim10K $\rightarrow$ Cityscapes scenario of \mname~models trained with different confidence settings. We visualise true positives in {\color{blue} blue} bbx, false negatives in {\color{red} red} bbx and false positives in {\color{orange} orange} bbx. By introducing a gradual transition between confidence metrics, we achieve less false positive detections compared to training only using $C_{det}$ and less false negative detections compared to training only using $C_{comb}$.}
  \label{fig:qualitative}
  \squeezeup
\end{figure*}

\subsection{Ablation study}
\label{sec:exp:ablation}

We verify the impact of the main design choices of our method with an ablation study on Sim10K $\rightarrow$ Cityscapes. 

\begin{figure}[t]
  \centering
  \includegraphics[width=1\linewidth]{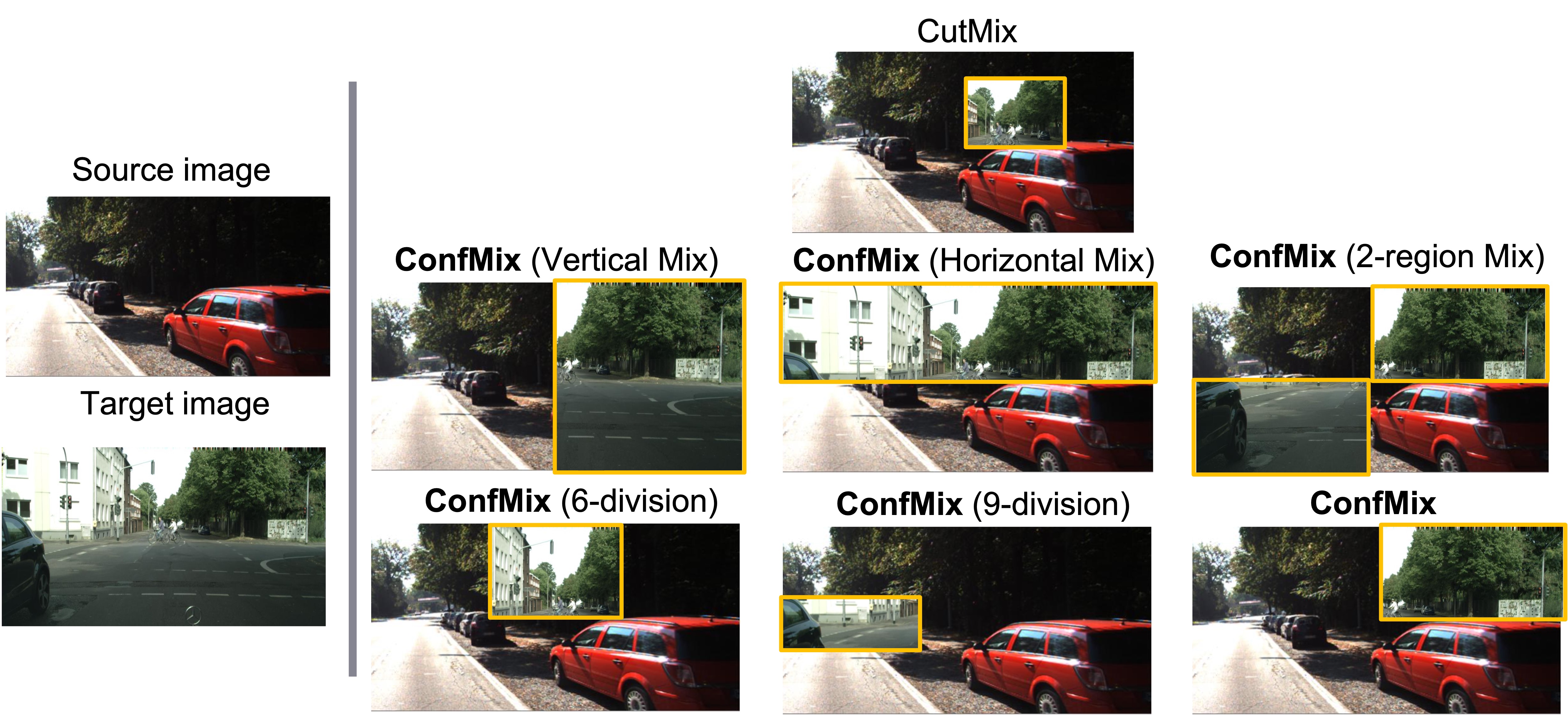}
  \caption{Illustration of different mixing strategies. \textbf{CutMix}~\cite{cutmix} randomly cuts a target region and mixes it with the source image. \mname~(Vertical Mix) vertically cuts the source and target image in the middle and mixes the most confident target region. \mname~(Horizontal Mix) horizontally cuts the source and target image in the middle and mixes the most confident target region. \mname~(2-region Mix) selects the two most confident regions of the target image for mixing. Finally, \mname~(6-division), \mname~(9-division), and \mname~divides the target image into 6, 9 and 4 regions, respectively, and selects only the most confident target region for mixing.}
  \label{fig:mixing_options}
  \squeezeup
\end{figure}

\noindent\textbf{Does the confidence-based region mixing help?} We analyse a variety of different mixing strategies and their impact in terms of object detection performance after adaptation. Specifically, we run our method using the $\ccname$ confidence
without the proposed progressive pseudo labelling. We vary the mixing strategy with 5 different options (as shown in Fig.~\ref{fig:mixing_options}):
\textbf{CutMix}~\cite{cutmix} randomly cuts a target region and mixes it with the source image; \mname~(Vertical Mix) vertically cuts both source and target images in the middle and mixes the most confident target region;
\mname~(Horizontal Mix) horizontally cuts both source and target images in the middle and mixes the most confident target region;
\mname~(2-region Mix) selects the two most confident regions of the target image for mixing; and \mname~selects only the most confident region of the target image for mixing. We further examine the 4-division scheme of \mname~by varying the number of divided regions into 6 ($2 \times 3$) and 9 ($3 \times 3$), respectively.

As can be seen in Table~\ref{tab:ablation_mixing}, our \mname~achieves a mAP gain of +5.6\% compared to CutMix, meaning that considering the most confident target region to mix is more beneficial than randomly cutting a target region and mixing it with the source image for adaptation.
\mname~shows also a superior performance than \mname~(Vertical Mix), \mname~(Horizontal Mix), and \mname~(2-region Mix). This means that mixing more target regions with the source image can impact the adaptation performance negatively, which might be due to the inclusion of a larger amount of less confident target pseudo detections. Interestingly, we notice that the cutting direction, i.e. vertical or horizontal, impacts the adaptation performance, where a vertical mixing demonstrates a better adaptation performance than the horizontal mixing; we believe that this phenomenon is scenario-dependent. In particular, for datasets concerning autonomous driving scenarios, a vertical mix always includes the road area of both target and source samples, thus being more likely to include objects, while a horizontal mix might only include further scenes where it is less likely to include objects.  {Finally, our 4-region division scheme leads to the best adaptation performance, with a mAP improvement of +0.6\% and +1.2\% compared to 6 and 9 divisions, respectively. This is supported by the intuition that smaller regions produce mixed samples containing a larger portion of the source domain, and increase the probability of occluded target objects, thus limiting the adaptive learning of their complete representation.}



\begin{table}[h]
\centering
\begin{tabular}{cc}
\hline
Method & mAP \\
\hline
CutMix~\cite{cutmix} & 49.1 \\
\mname~(Vertical Mix) & 53.6 \\
\mname~(Horizontal Mix) & 39.6 \\
\mname~(2-region Mix) & 41.1 \\
{\mname~(6-division)} & 54.1 \\
{\mname~(9-division)} & 53.5 \\
\hline
 \mname~ & \textbf{54.7} \\
\hline
\end{tabular}
\caption{Target detection accuracy with various mixing strategies.}
\label{tab:ablation_mixing}
\squeezeup
\end{table}


\noindent\textbf{Does the progressive pseudo labelling help?} 
We investigate a set of variants of pseudo labelling in order to verify the proposed progressive pseudo labelling strategy. Specifically, we ablate the usage of only $\caname$ or $\ccname$ for thresholding the pseudo detections. We also investigate different directions for the weight adjustment, i.e. $\caname$ $\rightarrow$ $\ccname$ and  $\ccname$ $\rightarrow$ $\caname$, as well as different shifting weights with $r$ (in Equation~\ref{eq:ratio}) representing a linear decay and $\delta$ (in Equation~\ref{eq:delta}) representing a non-linear decay. 
Our proposed strategy ($\caname$ $\rightarrow$ $\ccname$ ($\delta$)) gradually shifts from $\caname$ to the stricter $\ccname$ using the proposed shifting weight $\delta$.

As shown in Table~\ref{tab:ablation_conf}, using only $\ccname$ demonstrates to be more advantageous than using only $\caname$ for adaptation, and this is mainly due to more reliable pseudo detections. 
Moreover, we achieve the best result by gradually exploiting from $\caname$ to $\ccname$ with the non-linear weight $\delta$. A less restrictive confidence metric $\caname$ at the early adaptation allows more target pseudo detections and this can help with the target representation learning, while by gradually shifting to the usage of $\ccname$, we improve the reliability of pseudo detection, thus benefiting the detector accuracy. 

Figure~\ref{fig:qualitative} shows qualitative detection results when using only $\caname$, only $\ccname$, and our proposed strategy $\caname$ $\rightarrow$ $\ccname$ ($\delta$). 
As can be observed, the model is more likely to predict false positives when using only $\caname$, whereas the model using only $\ccname$ generates more false negatives. Our proposed gradual transition strategy can better combine the two confidence metrics for adaptive training, achieving the best adaptation performance on the target domain. 

\begin{table}[h]
\centering
\resizebox{\linewidth}{!}{%
\begin{tabular}{@{}lcccccc@{}}
\hline
Confidence &
  $\caname$ &
  $\ccname$ &
  $\ccname$ &
  $\caname$ &
  $\ccname$ &
  $\caname$ \\
   &
   &
   &
  $\rightarrow$ $\caname$ (r) &
  $\rightarrow$ $\ccname$ (r) &
  $\rightarrow$ $\caname$ ($\delta$) &
  $\rightarrow$ $\ccname$ ($\delta$) \\\hline
mAP &
  52.7 &
  54.7 &
  55.0 &
  54.9 &
  54.3 &
  \textbf{56.3} \\ \hline
\end{tabular}%
}
\caption{Target detection accuracy with various confidence-based pseudo labelling.}
\label{tab:ablation_conf}
\squeezeup
\end{table}



\noindent\textbf{Does the weight of consistency loss matter?} As pseudo detections are inevitably noisy, we are motivated to weight the consistency loss appropriately in order to avoid the introduction of pseudo detection errors. We therefore investigate how the weight of consistency loss impacts the adaptation performance by using a set of constant weights, in comparison to our dynamic weight $\gamma$. Regarding the constant weight, we vary it from 0.2 to 1 in the step of 0.2. As shown in Table~\ref{tab:ablation_weight}, using $\gamma$ as the consistency weight leads to the best mAP performance, with an improvement of +0.7\% compared to the best constant weight of 0.6. Therefore, the use of a dynamic weight whose value varies depending on the pseudo detections confidence, can improve the performance of the model by stabilising the training and mitigating the problem of over-fitting unreliable pseudo detections.


\begin{table}[h]
\centering
\begin{tabular}{ccccccc}
\hline
Weight & 0.2 & 0.4 & 0.6 & 0.8 & 1 & $\gamma$\\
\hline
mAP & 53.8 & 54.7 & 55.6 & 54.9 & 47.7  & \textbf{56.3}\\
\hline
\end{tabular}
\caption{Target detection accuracy with different weights on the consistency loss.}
\label{tab:ablation_weight}
\squeezeup
\end{table}


{
\noindent\textbf{Does the number of pseudo detections before mixing matter}? We analyse the number of pseudo detections before mixing by retaining with only 25\%, 50\%, 75\% and 100\% (i.e. our setting) of the most reliable pseudo detections in $\tilde{\textbf{y}}_T$, with the confidence threshold $C_{th}$ fixed for filtering the detections. 
As shown in Table~\ref{tab:ablation_num_pseudo_dets}, decreasing the number of pseudo detections consistently leads to a worse result than using all of them, i.e. our setting. Thanks to our progressive pseudo labelling scheme, most of the false positives could be filtered out already, thus detections with a confidence greater than $C_{th}$ are generally useful for the model to learn the target features.
}

\begin{table}[h]
\centering
{
\begin{tabular}{ccccc}
\hline
Pseudo detections (\%) & 25\% & 50\% & 75\% & 100\% \\
\hline
mAP & 45.1 & 48.9 & 51.7 & \textbf{56.3} \\
\hline
\end{tabular}}
\caption{{Target detection accuracy at varying number of pseudo detections.}}
\label{tab:ablation_num_pseudo_dets}
\squeezeup
\end{table}

\noindent\textbf{Does sample mixing performs better than simple self-training?}
We compare \mname~with a simple baseline that consists in applying naive finetuning with pseudo detections. With the model trained on the source domain, we generate pseudo detections on the target dataset using the same confidence threshold $C_{th}$ used in \mname~to filter the boxes on top of non-maximum suppression. We then expand the training dataset containing both source and target samples for further training.  
With such self-training, we achieve a mAP of 30.5 (Cityscapes$\rightarrow$Foggy Cityscapes), 55.4 (Sim10K$\rightarrow$Cityscapes) and 46.4 (KITTI$\rightarrow$ Cityscapes), which are much inferior than \mname~as reported in Table~\ref{tab:sk2c} and \ref{tab:c2f}, confirming the effectiveness of our proposal.


\section{Conclusion}
\label{sec:conclusion}
We proposed \mname, a novel confidence-based mixing method for adapting object detectors trained on a source domain to a target domain in an unsupervised manner. 
We introduced a region-level strategy for sample data augmentation by mixing the region of the target image with the most confident pseudo detections with the source image, and achieved adaptation with a consistency loss on the pseudo detections.
We also introduced the progressive pseudo labelling scheme by gradually restricting the confidence metric in order to facilitate the smooth transition from learning the target representation to improving detection accuracy.
We compared our approach with state-of-the-art methods, demonstrating its superior performance on two benchmarks. 
As future work, we will apply our method in other practical scenarios, other than autonomous driving, and improve its compatibility with different object detection frameworks.


\clearpage
\appendix
\section{Supplementary Material}
In this Supplementary Material, we provide additional experiments to demonstrate the effect of pretrained detector backbone, i.e. with/without COCO-pretrained weights, on the adaptation performance using our proposed \mname.
Moreover, we also justify key hyperparameter choices in the implementation, including the confidence threshold $C_{th}$ to filter the detections on top of non-maximum suppression, and $\alpha$ that is used to scale the impact of the current iteration to the total number of iterations for the calculation of the shifting weight $\delta$. Consistently with the main manuscript, we conduct the ablation study on the Sim10K $\rightarrow$ Cityscapes setup.

\noindent\textbf{Does backbone initialisation negate the effect of domain adaptation?} As there are works using ImageNet-pretrained networks \cite{Free_Lunch, Spatial_Attn, vs2022instance, Graph-induced, SCAN, Collaborative, zhou2022multi, li2022sigma, everypixelmatters} and works that do not specify whether they are pretrained or not \cite{SC-UDA, MeGA-CDA, cond_domain_norm, Seek_Sim}, we are motivated to examine how the initialisation of our backbone affects the proposed adaptation strategy. Therefore, we experiment with the initialisation of the backbone with random weights. Compared to the setting described in the main manuscript, we only vary the confidence threshold $C_{th}$ used to filter the detections on top of the non-maximum suppression from 0.25 to 0.3 for the random weight setting, in order to account for less reliable predictions at the initial training phase. As shown in Table \ref{tab:sk2c_preinit}, with random weights initialisation, Source only, \mname~and Oracle achieve lower performance than their corresponding ones with the COCO pretrained weights. 
Moreover, we notice that \mname~with random weights obtains a mAP gain of +12.3\% and +28.4\% compared to its Source-only counterpart, on Sim10K$\rightarrow$Cityscapes and KITTI$\rightarrow$Cityscapes, respectively. While with COCO-pretrained weights, \mname~achieves a mAP gain of +6.8\% and +12.3\% compared to its Source-only counterpart, on Sim10K$\rightarrow$Cityscapes and KITTI$\rightarrow$Cityscapes, respectively. This shows that the adaptation of \mname~is more effective when the backbone is not pretrained, although its general detection performance on the target domain is bounded by the Oracle's performance.

\noindent\textbf{How does $C_{th}$ for filtering out detections affect adaptation?} We experiment with varying confidence threshold $C_{th}$, i.e. 0.1, 0.25 (our setting), 0.5, and 0.7, to filter the detections on top of non-maximum suppression. 
The weight used to balance the consistency loss $\gamma=1$ is fixed throughout this experiment. 
As shown in Table \ref{tab:ablation_cth}, using $C_{th}=0.5$ leads to the best adaptation performance, with a mAP improvement of +10.6\% and +10.9\% compared to $C_{th}=0.1$ and $C_{th}=0.7$, respectively. While the use of low values for $C_{th}$ leads the model to keep erroneous pseudo detections, the use of high values filters out those pseudo detections that are useful for the model to learn the target features. 

However, the use of $C_{th}$ is closely related to $C_{th}^\gamma$, which is the confidence threshold used to calculate the reliability of pseudo detections $\gamma$, defined as the ratio of valid detections after non-maximum suppression with confidence greater than $C_{th}^\gamma$. 
We therefore vary $C_{th}^{\gamma}$ from 0.4 to 0.9 under $C_{th}=0.25$, and from 0.6 to 0.9 under $C_{th}=0.5$ (note that the value of $C_{th}^{\gamma}$ should be larger than $C_{th}$ to function meaningfully). From the results reported in Table~\ref{tab:ablation_cth_gamma}, we can see that the best performance is given by the combination of $C_{th}=0.25$ and $C_{th}^{\gamma}=0.5$, which is our experimental setting reported in the main manuscript.   

\begin{table}[t!]
\centering
\resizebox{\linewidth}{!}{
\begin{tabular}{cccccc}
\hline
 & & & & Sim10K$\rightarrow$ & KITTI$\rightarrow$  \\
 & & & & Cityscapes & Cityscapes \\
\hline
Method & Detector & Backbone & Pretrained & mAP & mAP \\
\hline
Source only & YOLOv5 & CSP-Darknet53 & No & 33.9 & 21.7 \\
\mname~(Ours) & YOLOv5 & CSP-Darknet53 & No & 46.2 & 50.1 \\
Oracle & YOLOv5 & CSP-Darknet53 & No & 64.1 & 64.1 \\
\hline
Source only & YOLOv5 & CSP-Darknet53 & COCO & 49.5 & 39.9 \\
\mname~(Ours) & YOLOv5 & CSP-Darknet53 & COCO & 56.3 & 52.2 \\
Oracle & YOLOv5 & CSP-Darknet53 & COCO & 70.3 & 70.3 \\
\hline
\end{tabular}
}
\caption{Quantitative results (mAP) for Sim10K/KITTI $\rightarrow$ Cityscapes benchmark.}
\label{tab:sk2c_preinit}
\end{table}

\begin{table}[t!]
\centering
\begin{tabular}{ccccc}
\hline
$C_{th}$ & 0.1 & 0.25 & 0.5 & 0.7 \\
\hline
mAP & 42.1 & 47.7 & \textbf{52.7} & 41.8 \\
\hline
\end{tabular}
\caption{Target detection accuracy with different confidence thresholds $C_{th}$.}
\label{tab:ablation_cth}
\end{table}

\begin{figure}[t!]
  \centering
  \includegraphics[width=\linewidth]{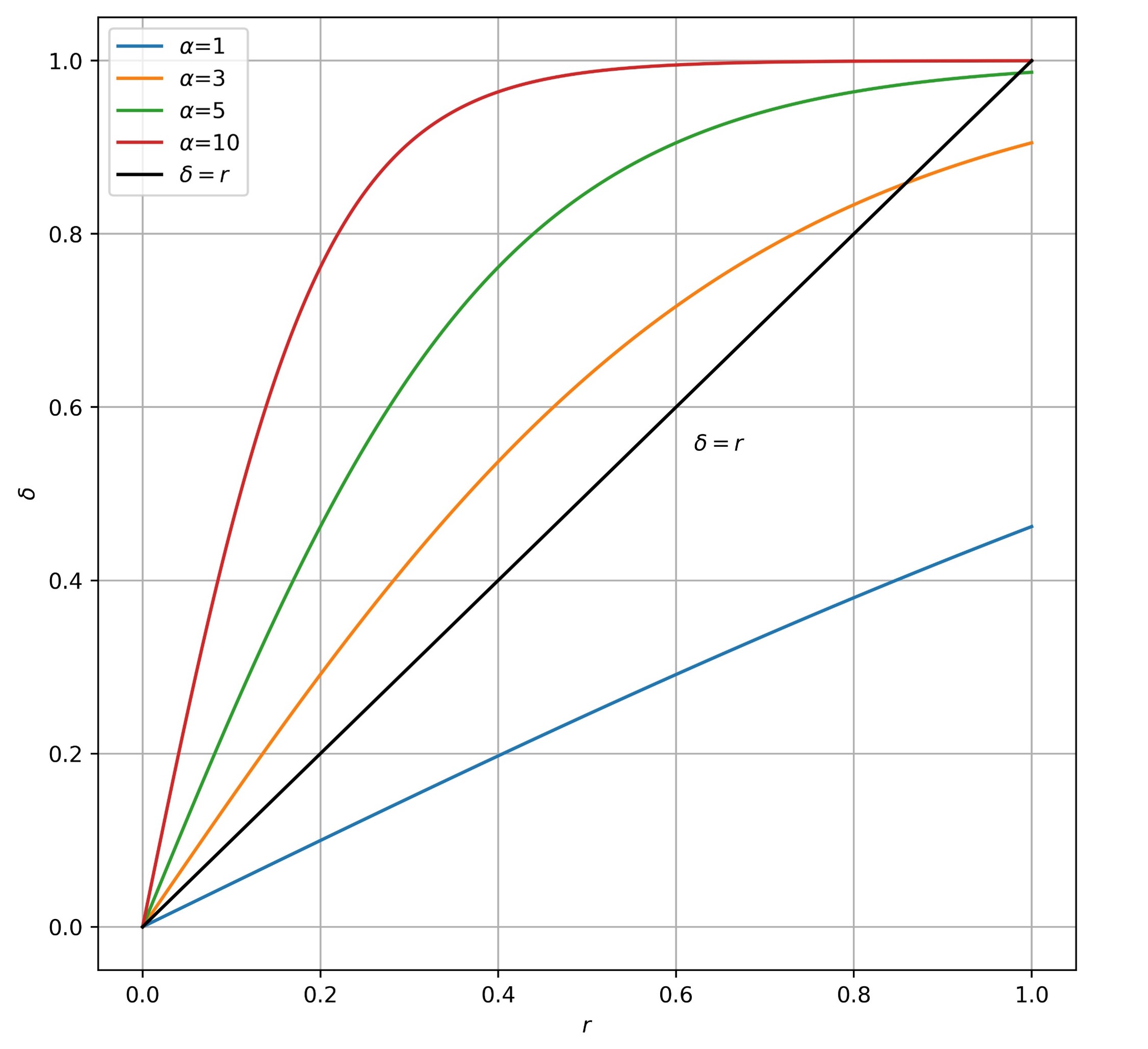}
  \caption{Visualisation of the evolution of $\delta$ throughout the training iterations with different values of $\alpha$. Note that $\delta=r$ is the linear function already ablated in the main manuscript.}
  \label{fig:vis_delta}
\end{figure}

\noindent\textbf{How does $\alpha$ for the confidence transition affect adaptation?} We analyse the effect of the hyperparameter $\alpha$ used to calculate the shifting weight $\delta$ for the smooth transition from $C_{det}$ to $C_{comb}$ during training. As illustrated in Figure \ref{fig:vis_delta}, a lower value of $\alpha$ gives a greater importance to the less strict confidence $C_{det}$, while a higher value of $\alpha$ gives a greater importance to the stricter confidence $C_{comb}$. As shown in Table \ref{tab:ablation_alpha}, using $\alpha=5$ leads to the best mAP performance, with an improvement of +0.8\% compared to $\alpha=1$, +0.6\% compared to $\alpha=3$, and +0.9\% compared to $\alpha=10$. 



\begin{table}[t!]
\centering
\resizebox{\linewidth}{!}{
\begin{tabular}{c|cccccc|cccc}
\hline
$C_{th}$ & 0.25 & 0.25 & 0.25 & 0.25 & 0.25 & 0.25 & 0.5 & 0.5 & 0.5 & 0.5\\ \hline
$C_{th}^{\gamma}$ & 0.4 & 0.5 & 0.6 & 0.7 & 0.8 & 0.9 & 0.6 & 0.7 & 0.8 & 0.9 \\\hline
mAP & 55.1 & \textbf{56.3} & 55.3 & 56.0 & 55.6 & 53.3 & 55.0 & 54.4 & 53.8 & 51.5\\
\hline
\end{tabular}
}
\caption{Target detection accuracy with different confidence thresholds $C_{th}^{\gamma}$ under $C_{th}=0.25$ and $C_{th}=0.5$.}
\label{tab:ablation_cth_gamma}
\end{table}

\begin{table}[t!]
\centering
\begin{tabular}{ccccc}
\hline
$\alpha$ & 1 & 3 & 5 & 10 \\
\hline
mAP & 55.5 & 55.7 & \textbf{56.3} & 55.4\\
\hline
\end{tabular}
\caption{Target detection accuracy with different confidence transition magnitudes.}
\label{tab:ablation_alpha}
\end{table}

{\small
\bibliographystyle{ieee_fullname}
\bibliography{egbib}
}

\end{document}